\documentclass{article} 
\usepackage{nips13submit_e,times}
\usepackage{url}
\usepackage{amsmath}
\usepackage{algorithm}
\usepackage{algorithmic}
\usepackage{graphicx}
\usepackage{hyperref}
\usepackage{verbatim}

\long\def\ignorethis#1{}

\newcommand{\vnorm}[1]{\left|\left|#1\right|\right|}

\newcommand{\policy}{\pi}
\newcommand{\return}{J}
\newcommand{\params}{\theta}
\newcommand{\reward}{r}
\newcommand{\state}{\mathbf{x}}
\newcommand{\action}{\mathbf{u}}

\newcommand{\velx}{v_x}
\newcommand{\posy}{p_y}

\renewenvironment{thebibliography}[1]{%
 \begin{oldthebibliography}{#1}%
   \setlength{\parskip}{0ex}%
   \setlength{\itemsep}{0ex}%
}%
{%
 \end{oldthebibliography}%
}

\title{Exploring Deep and Recurrent Architectures for Optimal Control}

\author{
Sergey Levine\\
Stanford University\\
\texttt{svlevine@cs.stanford.edu}
}

\nipsfinalcopy 

\begin{document}

\maketitle

\begin{abstract}
Sophisticated multilayer neural networks have achieved state of the art results on multiple supervised tasks. However, successful applications of such multilayer networks to control have so far been limited largely to the perception portion of the control pipeline. In this paper, we explore the application of deep and recurrent neural networks to a continuous, high-dimensional locomotion task, where the network is used to represent a control policy that maps the state of the system (represented by joint angles) directly to the torques at each joint. By using a recent reinforcement learning algorithm called guided policy search, we can successfully train neural network controllers with thousands of parameters, allowing us to compare a variety of architectures. We discuss the differences between the locomotion control task and previous supervised perception tasks, present experimental results comparing various architectures, and discuss future directions in the application of techniques from deep learning to the problem of optimal control.
\end{abstract}

\section{Introduction}

Multilayer neural networks such as autoencoders and convolutional neural networks have achieved state of the art results on a number of perception and natural language tasks \cite{dyda-cdptd-12,ksh-icdcn-12}\footnote{These are just a few recent examples, a good overview can be found in a recent survey paper \cite{bcv-rlrnp-13}.}. However, recent attempts to extend such methods from supervised (or semi-supervised) classification tasks to control or reinforcement learning have focused on adapting the ``deep'' component for perception, while using standard prior techniques for control, for example by learning a controller that operates directly on camera images \cite{lr-daenn-10,bdlvt-laptr-11}. While this is the most direct application of deep learning to optimal control, it leaves open the question of whether the control policy itself, rather than just the visual processing component, can benefit from deep learning techniques. In this paper, we explore this possibility by comparing several deep architectures, including recurrent and deep neural networks, on a continuous, high-dimensional locomotion task. In our experiments, the networks are used to represent the control policy, which maps joint angles to the torques applied at each joint.

A major stumbling block to the application of powerful models like deep neural networks to control has been a shortage of effective learning algorithms that can handle such rich policy classes while also addressing the sorts of complex tasks that can actually benefit from the increased representational power. Although the use of multilayer networks was explored in the eighties and nineties \cite{nw-ttbu-89,bayw-annlf-94}, such methods typically used small controllers for relatively simple tasks. Early experiments with neural network control represented both the system dynamics and policy as neural networks, so that the gradient of the policy could be propagated backwards in time \cite{nw-ttbu-89,gth-na-98}. However, this direct optimization approach can produce highly unstable gradients, and is often unsuitable for learning nontrivial behaviors. More recent reinforcement learning methods instead attempt to improve the policy by using sample rollouts on the actual system, which avoids the need to propagate gradients through time \cite{ps-rlmsp-08,kb-lmpr-09}. However, such methods are still susceptible to local optima, and work best with simple, linear policies, leading to the popularity of compact, specialized, and control-specific function approximators that lack the expressive power of large neural networks \cite{ins-lallm-03,ylv-ssblc-07}.

In this work, we apply a recently developed policy search algorithm that can effectively train large neural networks for difficult tasks by using trajectory optimization and example demonstrations. This algorithm, called guided policy search \cite{lk-gps-13}, uses trajectory optimization to ``guide'' the policy search to regions of high reward, typically initialized with a successful example demonstration. The policy is repeatedly optimized with a standard non-linear optimization algorithm, such as a LBFGS or SGD, making it relatively easy to combine the method with any controller class.

Using this algorithm, we will present a series of experiments that compare the performance of several deep and recurrent architectures on a continuous, high-dimensional locomotion task on rough terrain. This task differs qualitatively from previous perception tasks, since the input to the network consists of a relatively small number of joint angles, while the output consists of a comparable number of joint torques. This domain has much lower dimensionality than, for example, natural images, but also demands much greater precision, as locomotion and balance require precise feedbacks in response to continuous changes in joint angles. We evaluate shallow single-layer neural networks with soft and hard rectified linear units, as well as multi-layer networks with varying numbers of hidden units and recurrent networks that maintain a hidden state. Our preliminary results indicate a modest improvement in generalization for networks that are recurrent or have multiple layers, but also suggest that overfitting and local optima become a major problem for more sophisticated controller architectures. Improvements and more advanced experiments are discussed for future work.

\section{Why Deep Learning for Control?}

Reinforcement learning and optimal control has been applied with considerable success in fields such as robotics and computer graphics to control complex systems, such as robots and virtual characters \cite{ps-rlmsp-08,uypp-cancd-09}. However, many of the most successful applications have relied on policy classes that are either hand-engineered and domain-specific \cite{ks-pgrlf-04}, or else restricted to following a single trajectory \cite{ps-rlmsp-08}. Neither approach seems adequate for learning the sorts of rich motion repertoires that might needed, for example, for a robot that must execute a variety of tasks in a natural environment, or a virtual character that must exhibit a range of behaviors to emulate a human being. In order to scale up optimal control and reinforcement learning to such a wide range of tasks, we need general-purpose policies that can represent any behavior without extensive hand-engineering.

One way to accomplish this is by using more expressive policy classes. Empirical results in the domain of computer vision suggest that multilayer networks can learn hierarchical structures in the data, with more general, higher-level concepts captured in higher levels of the hierarchy \cite{lrmdc-bhlful-12}. Learning hierarchical structures in control and reinforcement learning has long been regonized as an important research goal \cite{bm-rahrl-03}, due to its potential to drastically improve generalization and transfer. In the context of motor control, generalization might include learning a skill (such as walking) in one context (such as on flat ground), and generalizing it without additional training to other contexts (such as uneven ground), which could be enabled by learning a commonly applicable control law (such as balance). While such a control law might not by itself accomplish walking, it can serve as one component in a hierarchically organized walking controller. In our evaluation, we examine a more modest example of generalization that can be attained with current methods.

Unfortunately, neural networks have also proven to be notoriously difficult to adapt for control of complex dynamical systems. In this work, we use the recently developed guided policy search algorithm to train neural network controllers. Compared to previous methods, this approach has been shown to be very effective for training neural network controllers \cite{lk-gps-13}, though it still struggles with local optima and does not appear to make the most of deep and recurrent architectures. We discuss possible solutions as future work.

\section{Guided Policy Search}

Guided policy search (GPS) is a reinforcement learning algorithm that uses trajectory optimization and example demonstrations to guide policy learning. The aim of the algorithm is to optimize the parameters $\params$ of a parameterized policy $\policy_\params(\action_t|\state_t)$, which specifies a distribution over actions $\action_t$ at each state $\state_t$. For a dynamical system such as a robot, the state might include features such as joint angles and velocities, while the actions might consist of the torques to apply at each joint. The desired task is defined by a reward function $\reward(\state_t,\action_t)$, and the goal of the learning algorithm is to maximize the expected return $E[\return(\params)] = E_{\policy_\params}[\sum_{t=1}^T \reward(\state_t,\action_t)]$, where $T$ is the time horizon. The guided policy search algorithm maximizes this objective by using an importance-sampled estimate of the expected return from off-policy samples. A rudimentary importance-sampled estimator is given by
\[
E[\return(\params)] \approx \frac{1}{Z(\params)}\sum_{i=1}^m \frac{\policy_\params(\state^i_1,\dots,\state^i_T,\action^i_1,\dots,\action^i_T)}{ q(\state^i_1,\dots,\state^i_T,\action^i_1,\dots,\action^i_T)} \sum_{t=1}^T \reward(\state^i_t,\action^i_t),
\]
The normalizing constant $Z(\params)$ ensures that the importance weights sum to 1, and the distribution $q$ denotes the sampling distribution. A more sophisticated regularized estimator is used in practice, and is discussed in previous work \cite{lk-gps-13}. By optimizing this estimate of the expected return with respect to the policy parameters, the policy can be improved.

When learning complex tasks such as legged locomotion, most trajectories receive very poor rewards (that is, all but a few torque patterns result in falling). Because of this, unless the sampling distribution $q$ is chosen carefully, all samples will receive a poor reward, and the optimization will be unable to improve the policy. The key idea in guided policy search is to draw the samples from a known good distribution over trajectories, which can be constructed by using a variant of the differential dynamic programming (DDP) trajectory optimization algorithm to construct a \emph{distribution} over trajectories (rather than a single trajectory). The simplest variant of GPS that is used in this paper begins with an example demonstration and then uses DDP to construct a distribution over similar trajectories that all have high reward. The algorithm first initializes the policy with supervised learning on samples drawn from this trajectory distribution, and then alternates between optimized the expected return and drawing additional on-policy samples to improve the return estimate. The optimization can be performed with any non-linear optimization algorithm. However, the importance weight normalization constant couples all of the samples, making it difficult to apply minibatch algorithms such as stochastic gradient descent (SGD). We therefore use either LBFGS or plain gradient descent in most of our experiments. A more recent variational variant of guided policy search eschews importance sampling and allows for any supervised learning algorithm, including SGD, to be used \cite{lk-vgps-13}. However, this variant currently only supports reactive policies that depend only on the current state, making it unsuitable for training recurrent neural networks. Extending variational GPS to handle recurrent neural networks would simplify the optimization problem.

\section{Locomotion on Rough Terrain}

\begin{figure}
\setlength{\unitlength}{0.5\columnwidth}
\begin{picture}(1.99,0.75) \linethickness{0.5pt}
\put(-0.05, 0.0){\includegraphics[height=0.12\columnwidth]{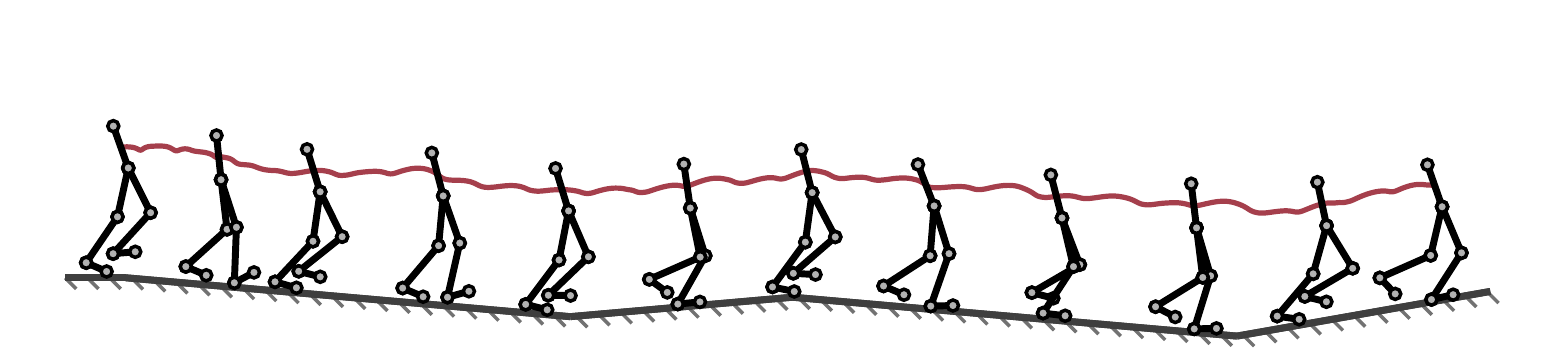}}
\put(-0.02, 0.15){\includegraphics[height=0.12\columnwidth]{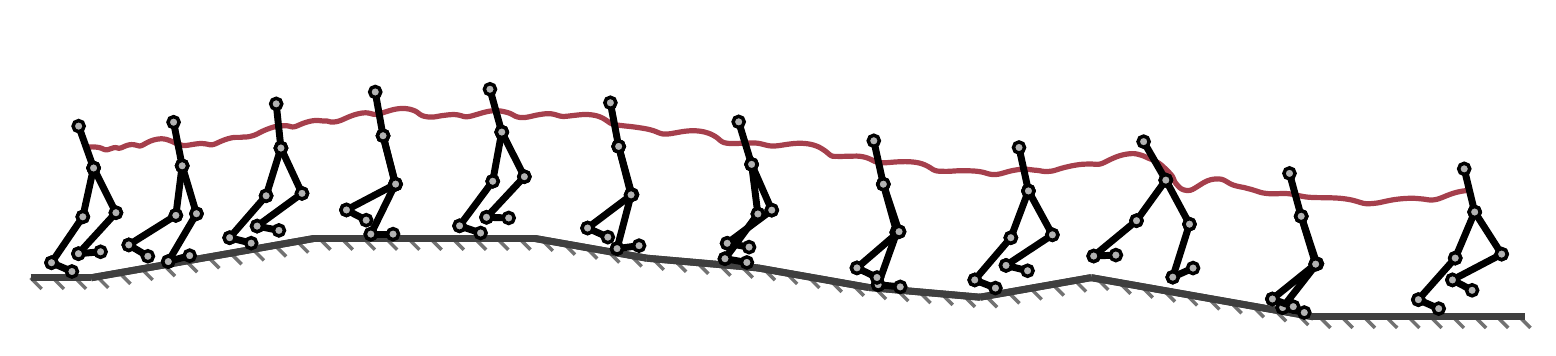}}
\put(-0.07, 0.3){\includegraphics[height=0.12\columnwidth]{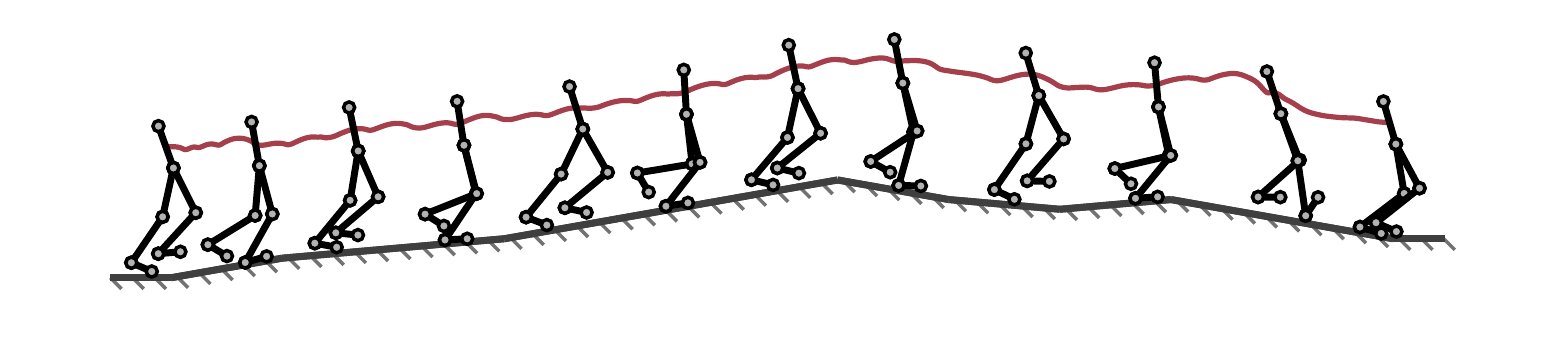}}
\put(-0.05, 0.45){\includegraphics[height=0.12\columnwidth]{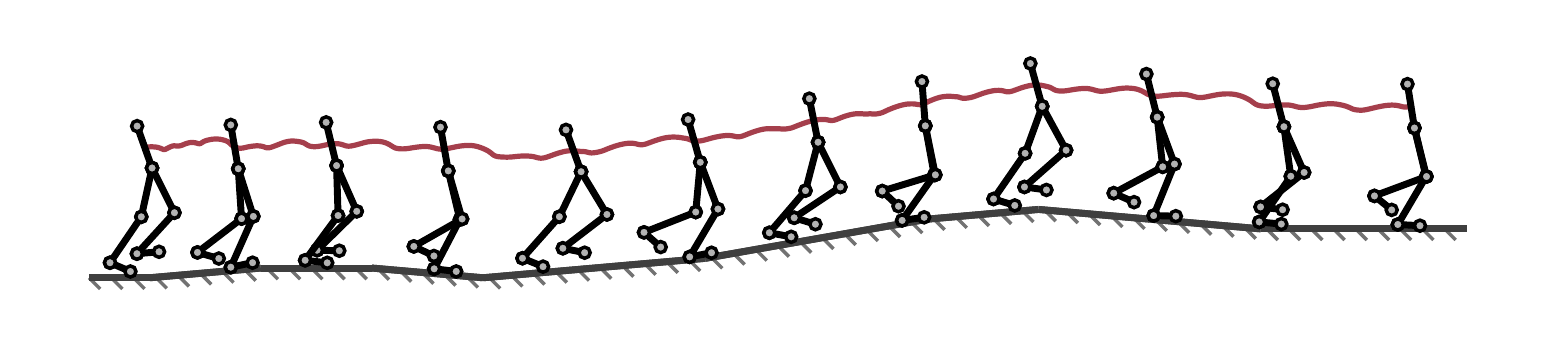}}
\put(-0.04, 0.6){\includegraphics[height=0.12\columnwidth]{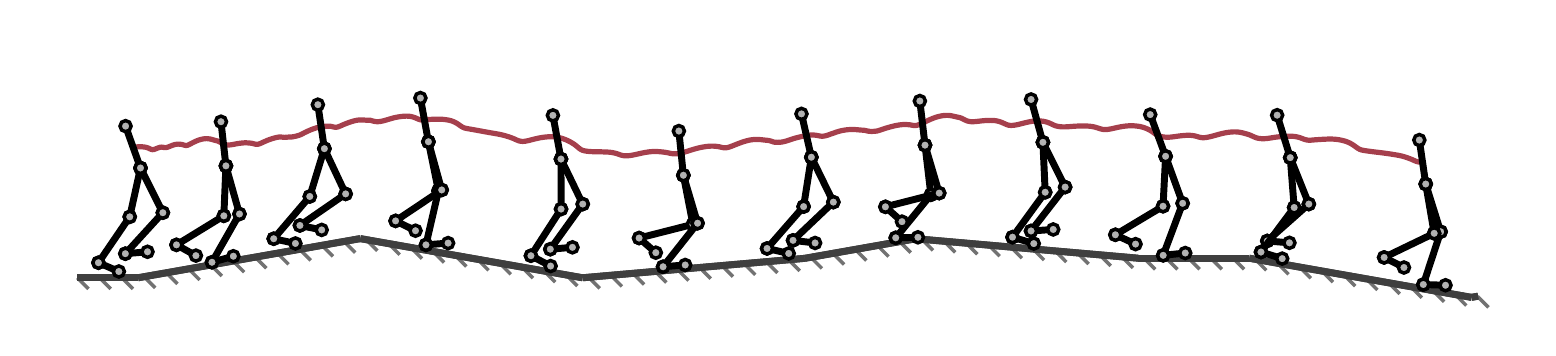}}
\put(0.34, 0.83){\footnotesize{Training Terrains}}

\put(1.015, 0.0){\includegraphics[height=0.12\columnwidth]{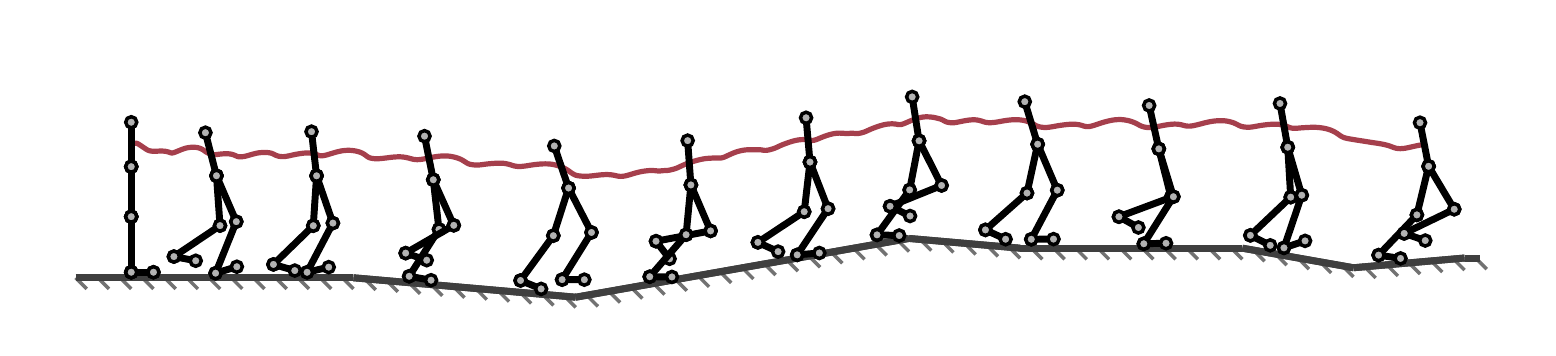}}
\put(1.015, 0.15){\includegraphics[height=0.12\columnwidth]{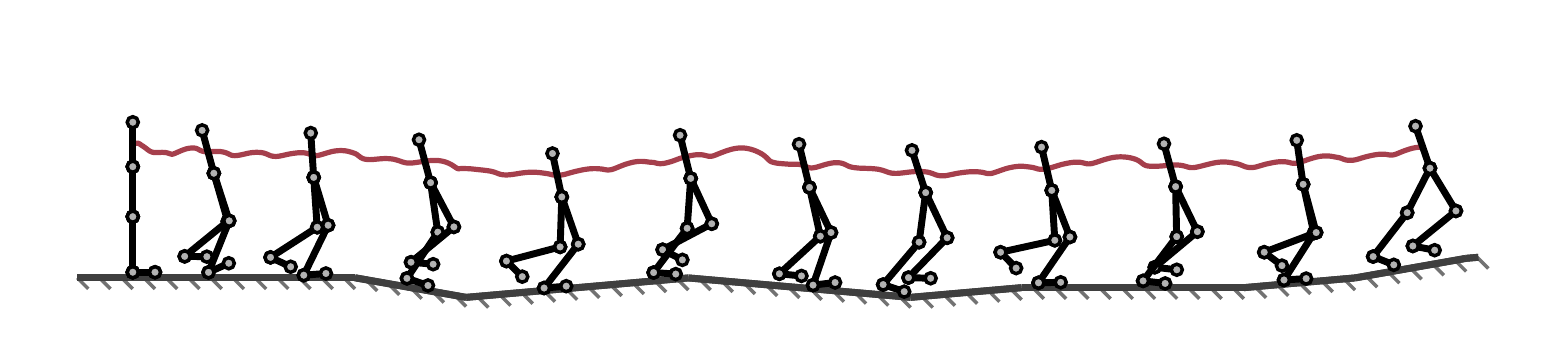}}
\put(0.99, 0.3){\includegraphics[height=0.12\columnwidth]{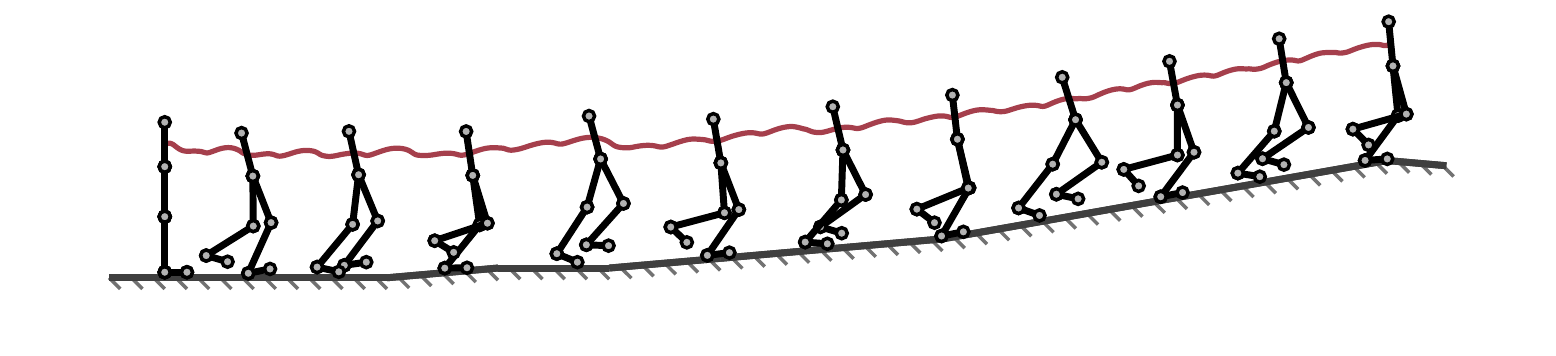}}
\put(1.0, 0.45){\includegraphics[height=0.12\columnwidth]{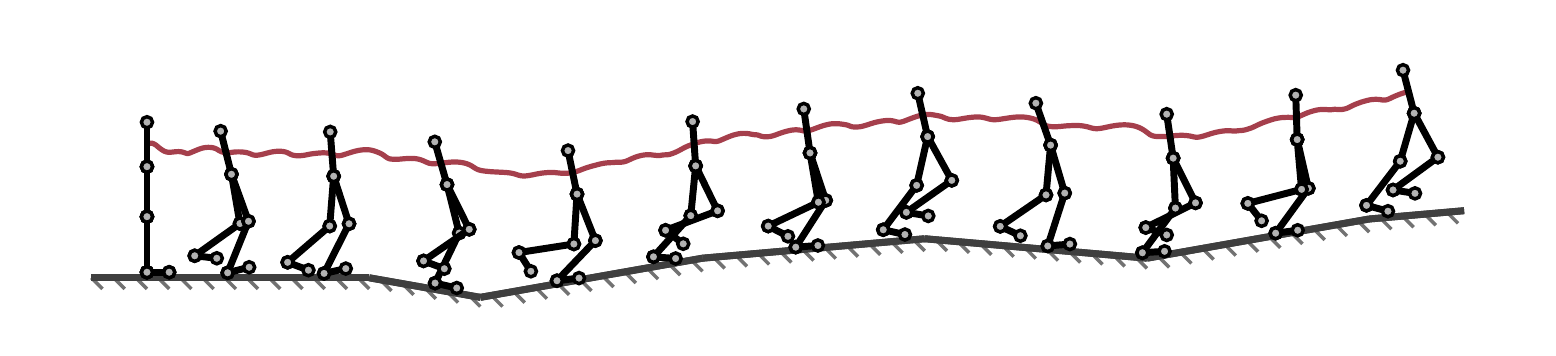}}
\put(0.98, 0.6){\includegraphics[height=0.12\columnwidth]{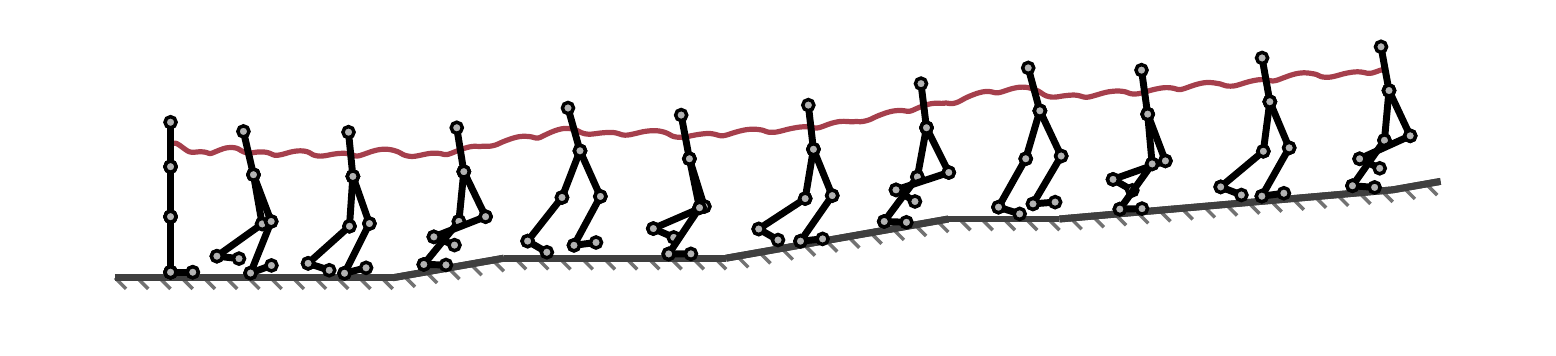}}
\put(1.38, 0.83){\footnotesize{Test Terrains}}
\end{picture}
\caption{Sections of five of the ten training terrains (left) and five of the ten test terrains (right), with overlaid rollouts of a recurrent neural network controller. The red lines indicate the trajectory of the trunk. All terrains consist of randomly chosen $0.5-1.0m$ segments with slopes between $-10^\circ$ and $+10^\circ$ (in $5^\circ$ increments).
\label{fig:domain}
}
\end{figure}

To evaluate a variety of neural network architectures, we chose a challening bipedal locomotion task. Since generalization is one of the most exciting advantages of richer policy representations, our evaluation is centered on testing the generalization of the learned policies. The task, visualized in Figure~\ref{fig:domain}, requires the policy to control a simulated planar biped walking on sloped terrain. The slope of the terrain varies from $-10^\circ$ to $+10^\circ$, and the reward function penalizes the squared deviation of the walker's velocity $\velx$ and trunk height $\posy$ from target values $\velx^\star = 1.2 m/s$ and $\posy^\star = 1.5 m$ (corresponding to a brisk jog), as well as the sum of squared torques:
\[
\reward(\state,\action) = -10^{-4}\vnorm{\action}^2 - (\velx - \velx^\star)^2 - 10(\posy - \posy^\star)^2.
\]
The control policy is trained on a varying number of example terrains, and tested on 10 different test terrains, as shown in Figure~\ref{fig:domain}. All terrains were generated randomly, but remain unchanged throughout the experiments. The learning algorithm was initialized with example demonstrations generated by a prior locomotion system \cite{ylv-ssblc-07}. In practice, a human demonstration or an offline planner might be used, as discussed in previous work \cite{lk-gps-13}. Note that although this task is modeled after the one in our previous work \cite{lk-gps-13}, the results are not directly comparable: the terrains are longer and more challenging, the simulator uses a more accurate contact model, the example demonstrations are different, and the motor noise was increased slightly. In short, the task was modified to be more difficult, so that differences between the various architectures would be more apparent.

To evaluate the performance of each controller on the test and training sets, we ran the controller for five trials on each terrain, and computed the fraction of total trials that were completed successfully -- that is, those trials where the controller did not fall and continued progressing forward. We believe that this metric is more informative than measuring the total accumulated reward, since a practical application would care much more about falling or failing to make progress than about the precise speed or torso height attained by the controller.

\section{Neural Network Controllers}

All controllers we evaluated used a neural network to map a vector of state features $\state_t$ to joint torques $\action_t$. The planar walker has 9 degrees of freedom, consisting of six joints, a global orientation, and a 2D global position. The state feature vector $\state_t$ omits the horizontal position, to prevent the controller from ``memorizing'' the pattern of slopes, and subtracts the ground height from the vertical position. The feature vector also includes the velocity at the root and all joints, an indicator for whether each foot is in contact, and the position of each body segment relative to the trunk (forward kinematics), for a total of $30$ dimensions.

It should be noted that this type of data is qualitatively different from image, audio, or video data that deep neural networks are more typically evaluated on. For example, compared to image data, the state feature vector is quite small. On the other hand, the specific numerical values in the vector are much more important, since the difference between a height of $1.0m$ and $0.5m$ could mean the difference between standing upright and falling to the ground. On the other hand, the specific brightness of a pixel often matters primarily in relation to other pixels, rather than in absolute terms. Similarly, the performance of the controller depends tremendously on the specific and very precise covariance between the inputs and outputs, since the controller must learn appropriate feedbacks to maintain balance. This type of output is significantly more complex than classification labels.

The subject of regularization is well studied in the field of deep learning. Some of the more popular regularizers use sparsity \cite{lnclp-oomdl-11} and denoising \cite{vlbm-ecrf-08} to avoid overfitting. While we did attempt to use a sparsity penalty in a few tests, we found that it only decreased the performance of the resulting controller. We believe that due to the differences in structure between image data and our state feature vectors, such regularizers would not be effective without modification. Developing regularization schemes that are more suited for optimal control is an interesting avenue for future work.

We evaluated three different types of network architectures: a single-layer network (referred to as shallow), a two-layer network (referred to as deep), and a single-layer recurrent network. For each network, we varied the number of hidden units, and used either soft or hard rectified linear units, given by $a = \log(1 + \exp(z))$ and $a = \max(0,z)$, respectively. Since we found that LBFGS struggled to optimize hard rectified linear units, we used gradient descent in those tests. While we did attempt to also evaluate networks with sigmoidal hidden units, we found that they performed very poorly. We suspect that sigmoidal units were unable to accurately represent the sorts of continuous feedbacks that are necessary for effective control policies.

\section{Results}

\begin{figure}
\setlength{\unitlength}{0.5\columnwidth}
\begin{picture}(1.99,0.88) \linethickness{0.5pt}
\put(1.45, 0.44){\includegraphics[height=0.22\columnwidth]{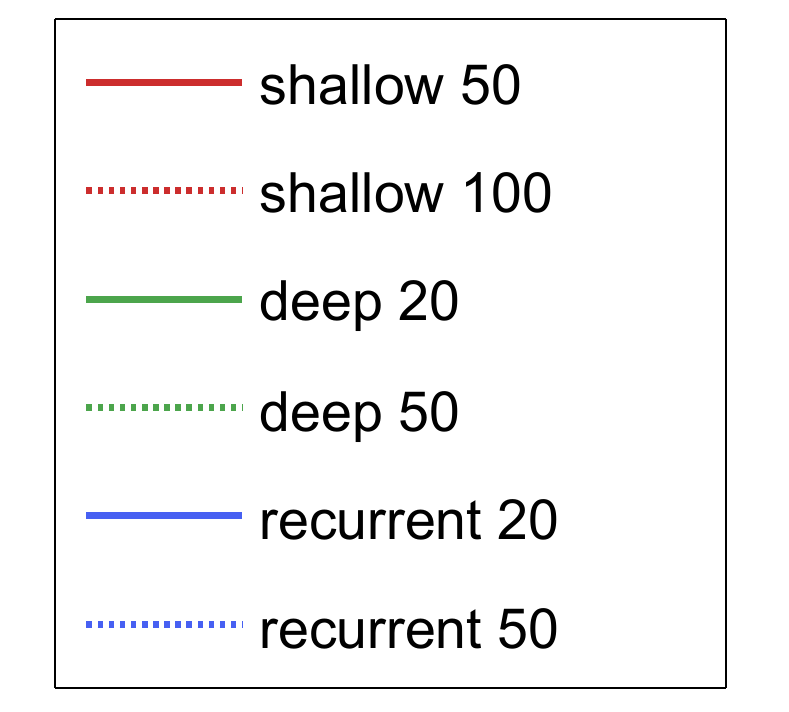}}

\put(0.75, 0.44){\includegraphics[height=0.22\columnwidth]{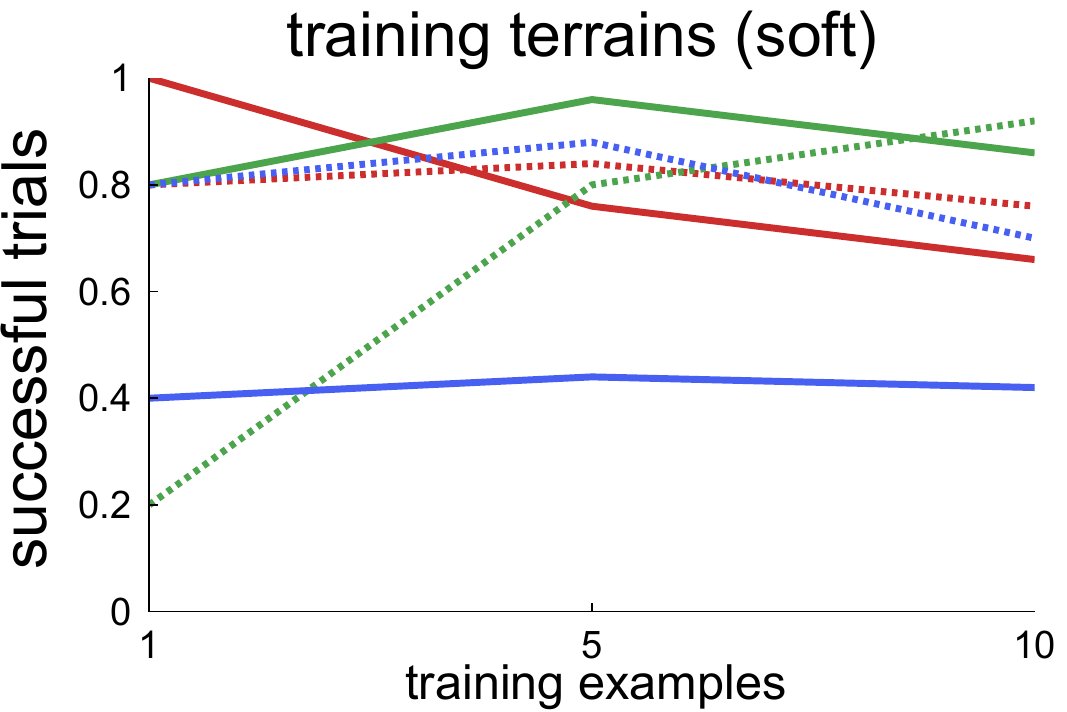}}
\put(0.05, 0.44){\includegraphics[height=0.22\columnwidth]{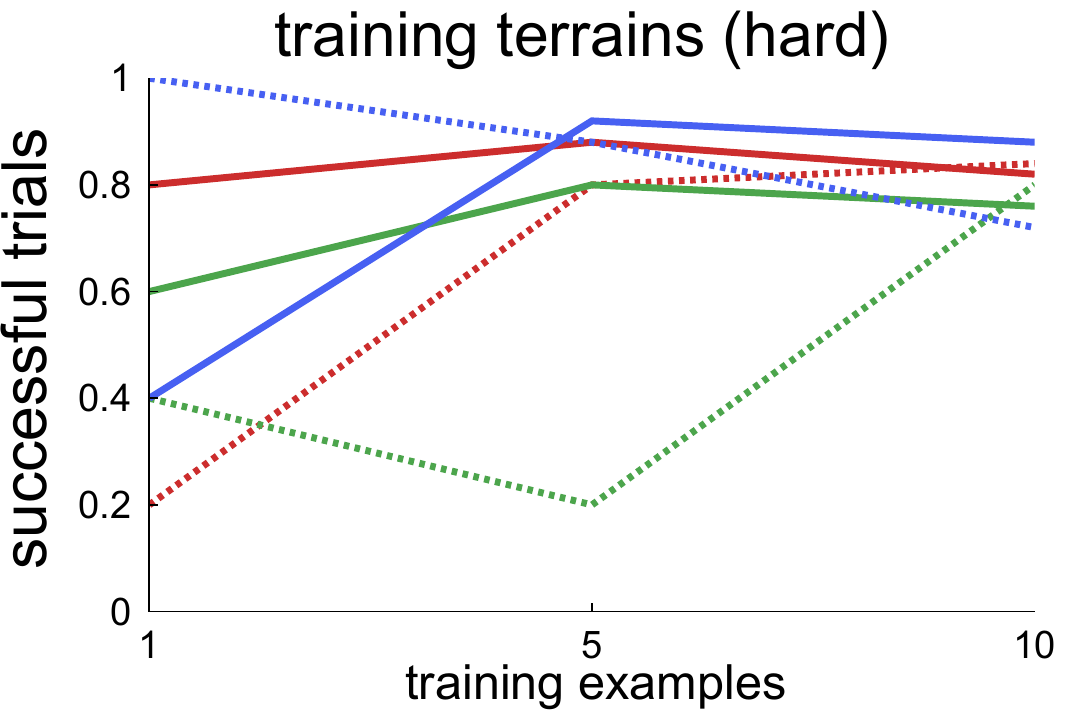}}

\put(0.75,-0.03){\includegraphics[height=0.22\columnwidth]{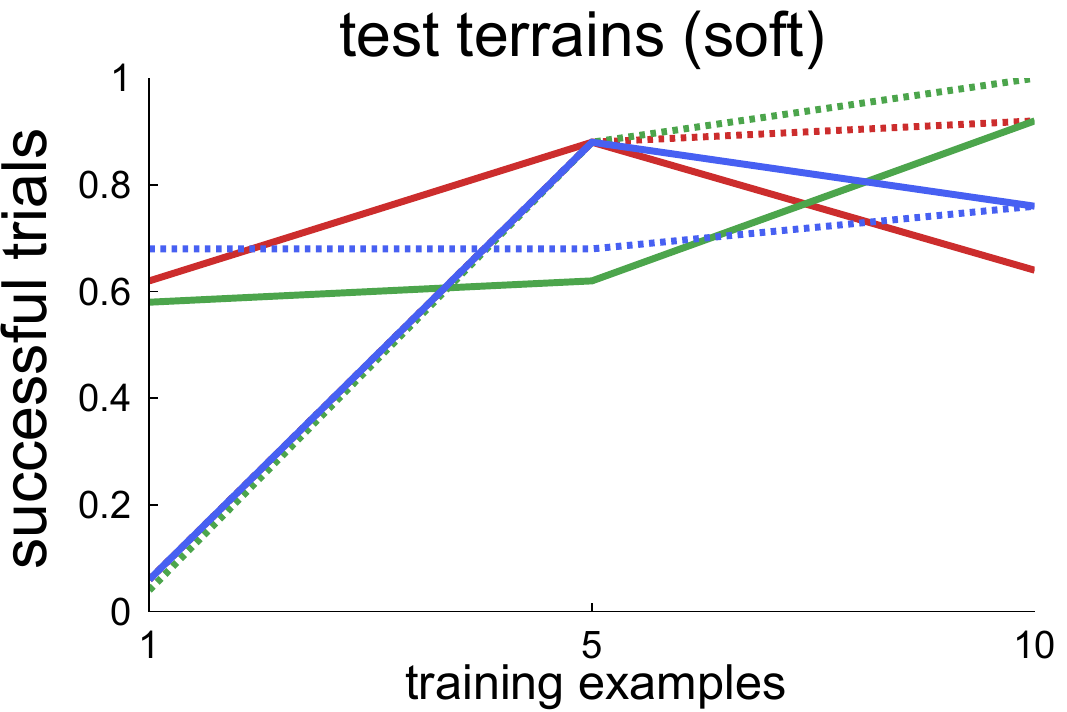}}
\put(0.05,-0.03){\includegraphics[height=0.22\columnwidth]{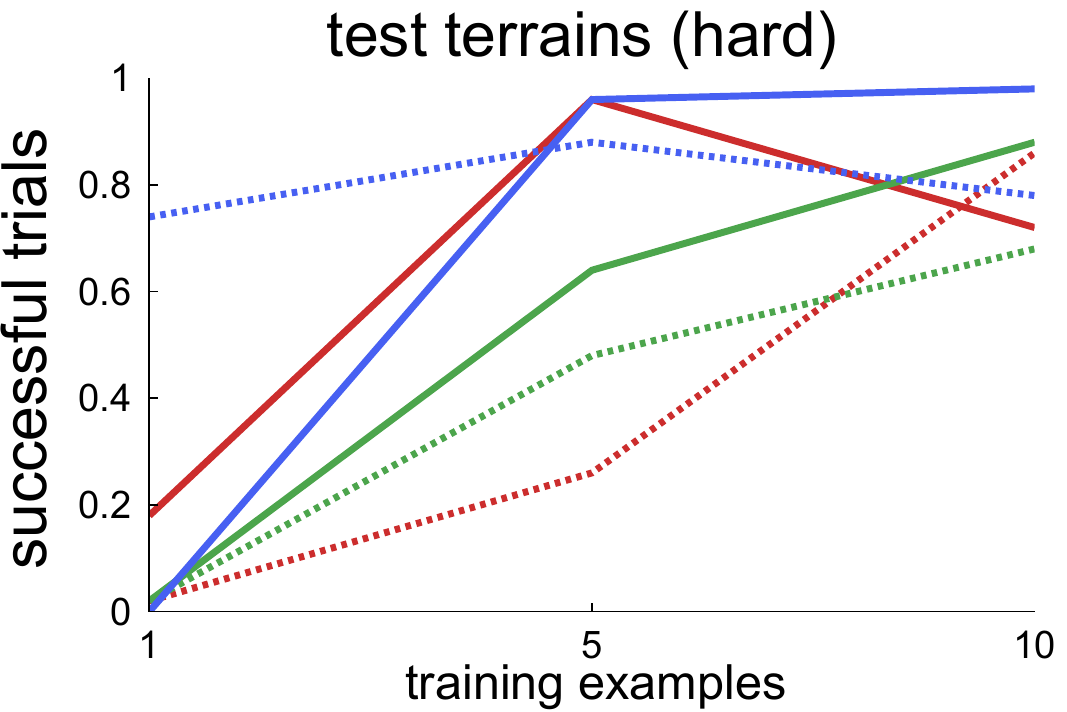}}
\end{picture}
\caption{Performance of each controller on its training set and the 10-terrain test set. Each terrain was traversed five times, producing slightly different results due to motor noise. The score of each controller is the fraction of trials that completed successfully (without falling or failing to make forward progress). Results for soft and hard units are shown separately for clarity.
\label{fig:results}
}
\end{figure}

The results of the rough terrain locomotion experiments are plotted in Figure~\ref{fig:results}. The legend lists the type of network (shallow, deep, or recurrent), followed by the number of hidden units in each layer. The number of hidden units for the shallow networks is chosen so that the number of parameters would roughly match the corresponding deep or recurrent network. The shallow 50-unit network has 1880 parameters, the 100-unit one has 3730, the 20-unit deep and recurrent networks have 1190, and the 50-unit ones have 4430. Results for soft and hard units are presented separately for clarity.

We trained each network on $1$, $5$, or $10$ example terrains. It should be noted that this number does not reflect the total number of \emph{samples} for the value estimate: $80$ samples are drawn for \emph{each} terrain from the example trajectory distribution, each consisting of $700$ time steps, and $10$ additional samples are drawn per terrain from the policy itself at each iteration of the GPS algorithm. Although we did not perform cross-validation, this would be a valuable addition in any future study, as it would help average out the effects of local optima and imbalances in the difficulty of the randomly generated terrains.

As expected, generalization improves with the number of training terrains, and more complex networks tend to require more training terrains to achieve good results. As the number of examples increases, deep and recurrent networks tend to outperform shallow ones, though not drastically, and not always. With soft rectified hidden units, the deep networks perform best, while the small recurrent network performs best when using hard rectified units. One possible explanation for this is that the second-order LBFGS algorithm (which was only used with soft units) can train deeper networks more effectively, while the piecewise linear hard units cause fewer problems for the long backpropagation through the recurrent networks, even without LBFGS. Aside from these more obvious trends, the results provide some interesting insights into the issues of overfitting and local optima, as they pertain to reinforcement learning and learning from demonstration.

\paragraph{Overfitting} Unlike with standard supervised learning, overfitting is a more complicated subject in this experiment, due to the combination of reinforcement learning and learning from demonstration. At a high level, controllers can suffer from two distinct forms of overfitting: they can overfit the example demonstration while learning a poor control law, resulting in poor performance on both training and test terrains, or they can overfit the particular pattern of slopes on the training terrain, resulting in poor performance on only the test terrain.

The results show both forms of overfitting. With hard hidden units, larger networks (shallow and deep) tend to overfit the example demonstrations, as shown by their poor performance on the training terrain with one example (this is also the case for the large deep network with soft units). This type of overfitting is characterized by an improvement in training terrain performance with more training examples, which would seem paradoxical in supervised learning.

On the other hand, the results also show a clear example of overfitting to the training terrain in the case of the 100-unit shallow network with hard hidden units and 5 examples. Surprisingly, this type of overfitting does not apear to severely afflict recurrent networks, which one might hypothesize would be the most vulnerable to ``memorizing'' the pattern of slopes in the training terrains, and instead appears to correspond simply to the size of the network, though it is unclear if this trend is real or coincidental. These problems could be alleviated with some sort of regularization, but as discussed previously, standard regularization methods are tricky to apply to such tasks.

\paragraph{Local Optima} By their nature, complex control tasks present a considerable number of local optima. This stems directly from the expected reward objective: objectives like maximum likelihood generally take an expectation under a fixed data distribution, while the control objective requires an expectation with respect to the distribution being optimized. This means that samples with low probability under the policy will not be represented in the objective value, regardless of their reward. While restarts can be used to alleviate this problem to some degree \cite{lk-gps-13}, it does appear to afflict many of the networks in our experiment as the number of examples (and therefore the number and variety of samples) increases.

This problem is characterized by a decrease in both training \emph{and} test terrain performance with more example terrains, and represents a troubling challenge to scaling neural network-based optimal control to more difficult problems. In essence, the challenge is that, even if the representational power of a network is sufficient to learn a good policy, the increasing number of poor local optima can make this impossible even if the number of examples is increased indefinitely. In the next section, we discuss future directions to address this challenge.

As an aside, pretraining has been viewed as one way to deal with local optima for deep networks \cite{bcv-rlrnp-13}. Multilayer networks can be pretrained as autoencoders, using some sort of sparsity or denoising for regularization. We did attempt to pretrain the deep networks with a sparsity regularizer, but found that this only harmed their performance. We suspect that this is related to the particular statistical properties of the state feature vectors, as mentioned previously in our discussion of regularizers.

\paragraph{Conclusions} Overall, the results suggest that, even for an equal number of parameters, deep and recurrent networks at least match, and sometimes outperform, shallow networks when the number of example terrains is large. This difference is not very prominent, though it is also likely that local optima and overfitting are hampering the larger deep and recurrent networks. The main lessons to be learned from this experiment are the following: in the context of control and learning from demonstration, overfitting can come in two flavors (overfitting demonstrations and overfitting example domains), local optima pose a serious challenge, especially for more complicated neural networks, and deep and recurrent networks do appear to offer some advantages over shallow ones, though the full benefit is likely not being realized. Lastly, a more practical lesson is that many of the larger networks only generalize well when the number of training domains is large. This is often missed in standard reinforcement learning experiments, which tend to be evaluated on one domain at a time.

\section{Future Work}

We presented experiments that compare the performance of various deep architectures on a continuous, high-dimensional locomotion task. Our analysis shows a modest improvement in generalization for both deep and recurrent networks when the number of training domains is large, though the results also indicate that such networks are likely hampered by local optima, which might pose a greater challenge in optimal control than in supervised learning.

These results point to a number of promising directions for future work. Since the local optima appear to be a bigger problem as the number of training domains increases, one possibility for addressing this challenge is to train controllers incrementally, starting with a few easy tasks and progressing to larger and larger numbers of training domains. However, it is unclear if a policy that falls into a poor local optima on the first few domains would ever recover if presented with additional ones. Another option is to explore regularization schemes that might alleviate such local optima. There is some hope that this is possible, since poor local optima are likely to correspond to ``unstable'' (in an informal sense) control laws. For example, a rough terrain could be traversed successfully by a controller that understands balance, but it could also be traversed by a controller that ``memorized'' a particular pattern of slopes. The latter controller is more brittle, and stochastic regularization schemes, like dropout \cite{hskss-innpc-12}, might be effective in avoiding such solutions.

Another promising direction is to modify the optimization algorithm to make it more like the tractable supervised learning objectives. A recent followup to the guided poliy search algorithm accomplishes just that, by using a variational approximation to a modified policy objective \cite{lk-vgps-13}. This algorithm alternates between optimizing a policy to match a trajectory distribution (in maximum likelihood terms), and optimizing the trajectory distribution to both achieve a high reward and match the policy. The difficulty with this approach is that the trajectory optimization step requires the policy to be differentiable in terms of the system state (which can be difficult if, as in our experiments, the policy uses non-differentiable features like contact indicators), and the optimization requires the policy to be reactive, preventing it from being used with recurrent networks. The former challenge could be addressed with a stochastic trajectory optimization algorithm, perhaps modeled on recent work on path integrals for optimal control \cite{tbs-rlmsh-10}, while the latter could be addressed by including the recurrent hidden unit state in the trajectory optimization.

A further advantage of switching to a maximum likelihood objective is that it would enable scalable optimization algorithm such as stochastic gradient descent to be used. This is likely to be a prerequisite for training on the large numbers of domains that would be required for more complex tasks. Even in our experiments, the average policy optimization run with 10 training terrains used $1800$ trajectories, each $700$ time steps in length, corresponding to $1.26$ million forward and backward propagations through the network for each gradient step.

\subsubsection*{Acknowledgments}

We thank the workshop reviewers for their constructive and helpful feedback. This work was supported by NSF Graduate Research Fellowship DGE-0645962 and an NVIDIA Graduate Fellowship.

\subsubsection*{References}
\vspace{-0.6cm}
\bibliographystyle{abbrv}
\bibliography{references}

\begin{thebibliography}{10}

\bibitem{bm-rahrl-03}
A.~G. Barto and S.~Mahadevan.
\newblock Recent advances in hierarchical reinforcement learning.
\newblock {\em Discrete Event Dynamic Systems}, 13(1-2):41--77, 2003.

\bibitem{bdlvt-laptr-11}
L.~Bazzani, N.~de~Freitas, H.~Larochelle, V.~Murino, and J.-A. Ting.
\newblock Learning attentional policies for tracking and recognition in video
  with deep networks.
\newblock In {\em International Conference on Machine Learning (ICML)}, pages
  937--944, 2011.

\bibitem{bayw-annlf-94}
F.~Beaufays, Y.~Abdel-Magid, and B.~Widrow.
\newblock Application of neural networks to load-frequency control in power
  systems.
\newblock {\em Neural Networks}, 7(1):183--194, 1994.

\bibitem{bcv-rlrnp-13}
Y.~Bengio, A.~Courville, and P.~Vincent.
\newblock Representation learning: A review and new perspectives.
\newblock {\em IEEE Transactions on Pattern Analysis and Machine Intelligence},
  35(8):1798--1828, 2013.

\bibitem{dyda-cdptd-12}
G.~E. Dahl, D.~Yu, L.~Deng, and A.~Acero.
\newblock Context-dependent pre-trained deep neural networks for
  large-vocabulary speech recognition.
\newblock {\em IEEE Transactions on Audio, Speech {\&} Language Processing},
  20(1):30--42, 2012.

\bibitem{gth-na-98}
R.~Grzeszczuk, D.~Terzopoulos, and G.~E. Hinton.
\newblock Neuroanimator: Fast neural network emulation and control of
  physics-based models.
\newblock In {\em ACM SIGGRAPH}, 1998.

\bibitem{hskss-innpc-12}
G.~E. Hinton, N.~Srivastava, A.~Krizhevsky, I.~Sutskever, and R.~Salakhutdinov.
\newblock Improving neural networks by preventing co-adaptation of feature
  detectors.
\newblock abs/1207.0580, 2012.

\bibitem{ins-lallm-03}
A.~Ijspeert, J.~Nakanishi, and S.~Schaal.
\newblock Learning attractor landscapes for learning motor primitives.
\newblock In {\em Advances in Neural Information Processing Systems (NIPS 16)},
  2003.

\bibitem{kb-lmpr-09}
J.~Kober and J.~Peters.
\newblock Learning motor primitives for robotics.
\newblock In {\em International Conference on Robotics and Automation}, 2009.

\bibitem{ks-pgrlf-04}
N.~Kohl and P.~Stone.
\newblock Policy gradient reinforcement learning for fast quadrupedal
  locomotion.
\newblock In {\em Proceedings of the {IEEE} International Conference on
  Robotics and Automation}, 2004.

\bibitem{ksh-icdcn-12}
A.~Krizhevsky, I.~Sutskever, and G.~E. Hinton.
\newblock Imagenet classification with deep convolutional neural networks.
\newblock In {\em Neural Information Processing Systems (NIPS)}, 2012.

\bibitem{lr-daenn-10}
S.~Lange and M.~Riedmiller.
\newblock Deep auto-encoder neural networks in reinforcement learning.
\newblock In {\em IJCNN}, pages 1--8, 2010.

\bibitem{lnclp-oomdl-11}
Q.~V. Le, J.~Ngiam, A.~Coates, A.~Lahiri, B.~Prochnow, and A.~Y. Ng.
\newblock On optimization methods for deep learning.
\newblock In {\em International Conference on Machine Learning (ICML)}, 2011.

\bibitem{lrmdc-bhlful-12}
Q.~V. Le, M.~Ranzato, R.~Monga, M.~Devin, G.~Corrado, K.~Chen, J.~Dean, and
  A.~Y. Ng.
\newblock Building high-level features using large scale unsupervised learning.
\newblock In {\em International Conference on Machine Learning (ICML)}, 2012.

\bibitem{lk-gps-13}
S.~Levine and V.~Koltun.
\newblock Guided policy search.
\newblock In {\em International Conference on Machine Learning (ICML)}, 2013.

\bibitem{lk-vgps-13}
S.~Levine and V.~Koltun.
\newblock Variational policy search via trajectory optimization.
\newblock In {\em Neural Information Processing Systems (NIPS)}, 2013.

\bibitem{uypp-cancd-09}
U.~Muico, Y.~Lee, J.~Popovic, and Z.~Popovic.
\newblock Contact-aware nonlinear control of dynamic characters.
\newblock {\em ACM Transactions on Graphics}, 28(3), 2009.

\bibitem{nw-ttbu-89}
D.~Nguyen and B.~Widrow.
\newblock The truck backer-upper: an example of self-learning in neural
  networks.
\newblock In {\em Proceedings of the International Joint Conference on Neural
  Networks}, 1989.

\bibitem{ps-rlmsp-08}
J.~Peters and S.~Schaal.
\newblock Reinforcement learning of motor skills with policy gradients.
\newblock {\em Neural Networks}, 21(4):682--697, 2008.

\bibitem{tbs-rlmsh-10}
E.~Theodorou, J.~Buchli, and S.~Schaal.
\newblock Reinforcement learning of motor skills in high dimensions: a path
  integral approach.
\newblock In {\em ICRA}, 2010.

\bibitem{vlbm-ecrf-08}
P.~Vincent, H.~Larochelle, Y.~Bengio, and P.-A. Manzagol.
\newblock Extracting and composing robust features with denoising autoencoders.
\newblock In {\em International Conference on Machine Learning (ICML)}, 2008.

\bibitem{ylv-ssblc-07}
K.~Yin, K.~Loken, and M.~van~de Panne.
\newblock {SIMBICON}: simple biped locomotion control.
\newblock {\em ACM Transactions Graphics}, 26(3), 2007.

\end{thebibliography}

\end{document}